\def\year{2020}\relax
\documentclass[letterpaper]{article} % DO NOT CHANGE THIS
\usepackage{aaai20}  % DO NOT CHANGE THIS
\usepackage{times}  % DO NOT CHANGE THIS
\usepackage{helvet} % DO NOT CHANGE THIS
\usepackage{courier}  % DO NOT CHANGE THIS
\usepackage[hyphens]{url}  % DO NOT CHANGE THIS
\usepackage{graphicx} % DO NOT CHANGE THIS
\usepackage{graphicx}  % DO NOT CHANGE THIS
\usepackage[small]{caption}
\usepackage{algorithm}
\usepackage{algorithmic }
\usepackage{multicol}
\usepackage{amsmath}
\usepackage{textcomp}
\usepackage{mathtools}
\usepackage{array}
\usepackage{wrapfig}
\usepackage{amsmath, amsthm, amssymb, amsfonts}
\usepackage{xcolor}
\usepackage{comment}
\usepackage{flushend}

\title{Learning and Reasoning for Robot Sequential Decision Making under Uncertainty}
\author{Saeid Amiri$^1$, Mohammad Shokrolah Shirazi$^2$, Shiqi Zhang$^1$\\
{$^1$ SUNY Binghamton, Binghamton, NY 13902 USA}\\
{$^2$ University of Indianapolis, Indianapolis, IN 44227 USA}\\
\texttt{samiri1@binghamton.edu; shirazim@uindy.edu; szhang@cs.binghamton.edu}
}

\begin{document}

\maketitle

\begin{abstract}
Robots frequently face complex tasks that require more than one action, where sequential decision-making (\textsc{sdm}) capabilities become necessary. 
The key contribution of this work is a robot \textsc{sdm} framework, called \textsc{lcorpp}, that supports the simultaneous capabilities of supervised learning for passive state estimation, automated reasoning with declarative human knowledge, and planning under uncertainty toward achieving long-term goals. 
In particular, we use a hybrid reasoning paradigm to refine the state estimator, and provide informative priors for the probabilistic planner. 
In experiments, a mobile robot is tasked with estimating human intentions using their motion trajectories, declarative contextual knowledge, and human-robot interaction (dialog-based and motion-based). 
Results suggest that, in efficiency and accuracy, our framework performs better than its no-learning and no-reasoning counterparts in office environment.
\end{abstract}

\section{Introduction}
\label{sec:intro}
Mobile robots have been able to operate in everyday environments over extended periods of time, and travel long distances that have been impossible before, while providing services, such as escorting, guidance, and delivery~\cite{hawes2017strands,veloso2018increasingly,khandelwal2017bwibots}. 
Sequential decision-making (\textsc{sdm}) plays a key role toward robot long-term autonomy, because real-world domains are stochastic, and a robot must repeatedly estimate the current world state and decide what to do next.

There are at least three AI paradigms, namely \textit{supervised learning}, \textit{automated reasoning}, and \textit{probabilistic planning}, that can be used for robot \textsc{sdm}. 
Each of the three paradigms has a long history with rich literature. 
\textit{However, none of the three completely meet the requirements in the context of robot \textsc{sdm}. }
First, a robot can use supervised learning to make decisions, e.g., to learn from the demonstrations of people or other agents~\cite{argall2009survey}. 
However, the methods are not designed for reasoning with declarative contextual knowledge that are widely available in practice. 
Second, knowledge representation and reasoning (\textsc{krr}) methods can be used for decision making~\cite{gelfond2014knowledge}. 
% e.g., people show up in open-house events need guidance help, and students leaving classrooms do not. 
However, such knowledge can hardly be comprehensive in practice, and robots frequently find it difficult to survive from inaccurate or outdated knowledge. 
Third, probabilistic planning methods support active information collection for goal achievement, e.g., using decision-theoretic frameworks such as Markov decision processes (\textsc{mdp}s)~\cite{puterman2014markov} and partially observable \textsc{mdp}s (\textsc{pomdp}s)~\cite{kaelbling1998planning}. 
However, the planning frameworks are ill-equipped for incorporating declarative contextual knowledge. 

% We develop a robot SDM framework in this work, where robots are able to simultaneously learn from past experiences, reason with declarative contextual knowledge, and plan to achieve long-term goals under uncertainty. 
% In this work, SDM is on whether to interact with people, and (if so), when and how. 
% Since any of the mentioned methods have their pros and cons, we decide to integrate all three of them so that each method's weakness is resolved by the other ones. 
% With learning and reasoning, the robot uses offline learning while in sequential decision making, it can actively collect information. In section \ref{sec: RW}, we mention some works that have integrated some of these techniques to improve robot's desired performance. 

\begin{figure*}[t]
  \begin{center}
    \vspace{-1em}
     \includegraphics[width=1.65\columnwidth]{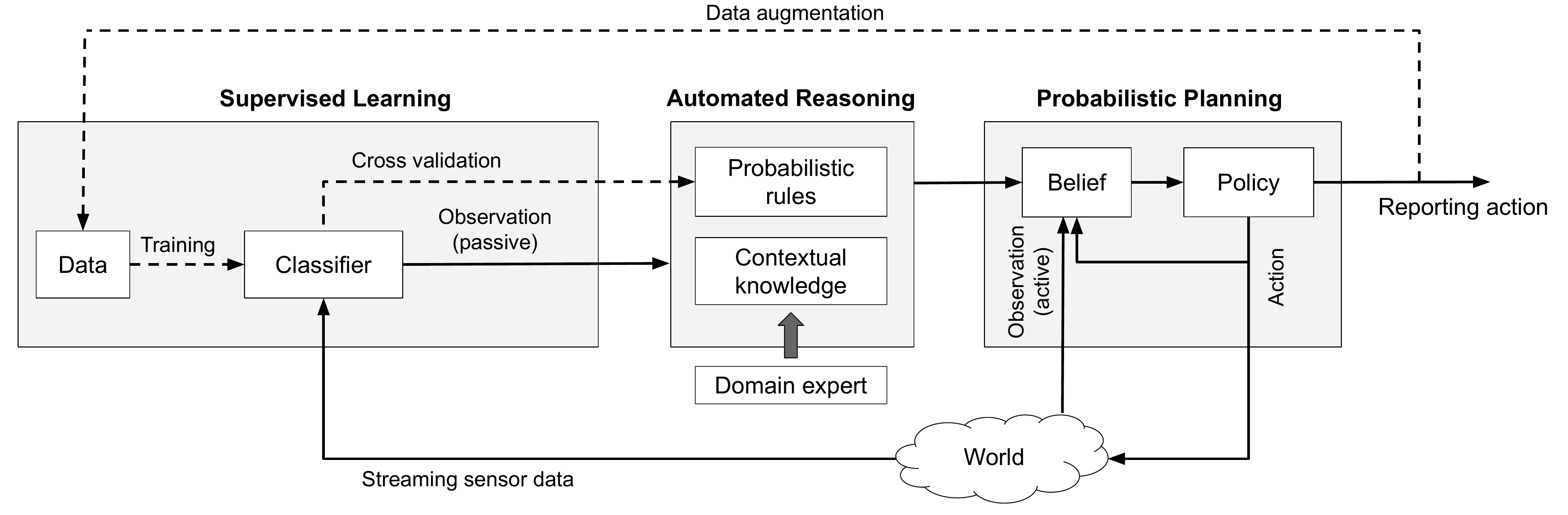}
    \vspace{-1em}
    \caption{An overview of \textsc{lcorpp} that integrates supervised learning, automated reasoning, and probabilistic planning. Streaming sensor data, e.g., from \textsc{rgb-d} sensors, is fed into an offline-trained classifier. 
The classifier's output is provided to the reasoner along with classifier's cross-validation. This allows the reasoner to encode uncertainty of the classifier's output.  
The reasoner reasons with declarative contextual knowledge provided by the domain expert, along with the classifier's output and accuracies in the form of probabilistic rules, and produces a prior belief distribution for the probabilistic planner. 
The planner suggests actions to enable the robot to actively interact with the world, and determines what actions to take, including when to terminate the interaction and what to report. At each trial, the robot collects the information gathered from the interaction with the world. In case of limited experience for training, \textsc{lcorpp} supports data augmentation through actively seeking human supervision. 
 Solid and dashed lines correspond to Algorithms~\ref{alg:lcorpp} and~\ref{alg:dt-control} respectively, as detailed in Section~\ref{sec:framework}. 
    }
    \label{fig:overview}
  \end{center}
  \vspace{-1.5em}
\end{figure*}

% \noindent
In this work, we develop a robot \textsc{sdm} framework that enables the simultaneous capabilities of learning from past experiences, reasoning with declarative contextual knowledge, and planning toward achieving long-term goals. 
Specifically, we use \emph{long short-term memory} (\textsc{lstm})~\cite{hochreiter1997long} to learn a classifier for passive state estimation using streaming sensor data, and use \emph{commonsense reasoning and probabilistic planning} (\textsc{corpp})~\cite{zhang2015corpp} for active perception and task completions using contextual knowledge and human-robot interaction. 
Moreover, the dataset needed for supervised learning can be augmented through the experience of active human-robot communication, which identifies the second contribution of this work.  
The resulting algorithm is called learning-\textsc{corpp} (\textsc{lcorpp}), as overviewed in Figure~\ref{fig:overview}.

We apply \textsc{lcorpp} to the problem of \emph{human intention estimation} using a mobile robot. 
The robot can passively estimate human intentions based on their motion trajectories, reason with contextual knowledge to improve the estimation, and actively confirm the estimation through human-robot interaction (dialog-based and motion-based). 
% The objective is to determine human intention (e.g., human intending to interact with the robot or not), accurately and promptly. 
% It should be noted that human intention may change over time, where the robot wants to adjust its estimation at runtime and only interact with those who are interested in interacting with the robot. 
Results suggest that, in comparison to competitive  baselines from the literature, \textsc{lcorpp} significantly improves a robot's performance in state estimation (in our case, human intention) in both accuracy and efficiency. \footnote{A demo video is available: https://youtu.be/YgiB1fpJgmo}

% This work is related to existing research that incorporates knowledge representation and reasoning (\textsc{krr}) into sequential decision-making (\textsc{sdm}) in stochastic worlds. 
% \textsc{sdm} can be realized via either probabilistic planning (e.g., \textsc{mdp}s and \textsc{pomdp}s~\cite{kaelbling1998planning}) or reinforcement learning (\textsc{rl})~\cite{sutton2018reinforcement}. 

\section{Related Work}
\label{sec:related}

\paragraph{\textsc{krr} and \textsc{sdm}:}
\textsc{sdm} algorithms can be grouped into two classes depending on the availability of the world model, namely probabilistic planning, and reinforcement learning (\textsc{rl}). 
Next, we select a sample of the \textsc{sdm} algorithms, and make comparisons with \textsc{lcorpp}. 

%%%%%%%%%%%%%Knowledge and RL/Planning %%%%%%%%%%%%%%%%
When world model is unavailable, \textsc{sdm} can be realized using \textsc{rl}. People have developed algorithms for integrating \textsc{krr} and \textsc{rl} methods~\cite{leonetti2016synthesis,yang2018peorl,lyu2019sdrl,sridharan2014integrating,griffith2013policy,illanes2019symbolic}. 
Among them, \citeauthor{leonetti2016synthesis} used declarative action knowledge to help an agent to select only the reasonable actions in \textsc{rl} exploration. 
\citeauthor{yang2018peorl} developed an algorithm called \textsc{peorl} that integrates hierarchical \textsc{rl} with task planning, and this idea was applied to deep \textsc{rl} by~\citeauthor{lyu2019sdrl} in \citeyear{lyu2019sdrl}. 

% These works are not benefiting from the available labeled data that can be used by the robot. 
%In that work, \textsc{rl} (low-level) helps learn action costs for the task planner, and the task planner %guides \textsc{rl} (high-level) to accomplish complex tasks.
% Also, researchers proposed an architecture that uses knowledge to find out \textsc{rl} components for incremental learning of unknown rules governing the domain dynamics \cite{sridharan2014integrating}. Symbolic planning has been incorporated to deep reinforcement learning to improve interpretability of subtasks \cite{lyu2019sdrl}.
% The work by \citeauthor{griffith2013policy} integrates human feedback for policy shaping in their reinforcement learning framework \cite{griffith2013policy}. 
% These works cannot learn complex representations from previous annotated decision-making experiences. 
%In case of world model being available, probabilistic planning methods can be used for computing action policies. 

When world models are provided beforehand, one can use probabilistic planning methods for \textsc{sdm}. 
Contextual knowledge and declarative reasoning have been used to help estimate the current world state in probabilistic planning~\cite{zhang2015mixed,zhang2015corpp,sridharan2019reba,ferreira2017answer,chitnis2018integrating,lu2018robot,lu2017leveraging}. 
Work closest to this research is the \textsc{corpp} algorithm, where hybrid reasoning (both logical and probabilistic) was used to guide probabilistic planning by calculating a probability for each possible state~\cite{zhang2015corpp}. 
% We use \textsc{corpp} in this work. 
% More recently, hybrid reasoning has been used to reason about world dynamics~\cite{zhang2017dynamically}, enabling planners to generate behaviors that are adaptive to dynamic world dynamics. 
% \cite{sridharan2019reba} developed a refinement-based architecture, where declarative knowledge is used for task planning and reasoning tasks, such as diagnosis and history explanation, and high-level deterministic plans are implemented via probabilistic planners. 
% Also, researchers have studied non-stationary Markov decision processes in which policies can be updated when there is change in the world states \cite{ferreira2017answer}.  
More recently, researchers have used human-provided declarative information to improve robot probabilistic planning~\cite{chitnis2018integrating}. 
% In a new study, researchers introduced a framework in which the agent learns world state transitions via model-based \textsc{rl}, and then incorporates them into the logical-probabilistic reasoning module \cite{lu2018robot}. 
% Also, the work by Lu et al. incorporates multi-modal perception and commonsense reasoning on a \textsc{pomdp}-based dialog manager to better find human intentions through engaging in a dialogue \cite{lu2017leveraging}. However, they are not leveraging any learning from datasets and do not consider the human-robot interaction initiation. 

% The main difference from the algorithms is that \textsc{lcorpp} is able to simultaneously learn from datasets as well as recently labeled-experiences. 

The main difference from the algorithms is that \textsc{lcorpp} is able to leverage the extensive data of (labeled) decision-making experiences for continuous passive state estimation. 
In addition, \textsc{lcorpp} supports collecting more data from the human-robot interaction experiences to further improve the passive state estimator over time.

\vspace{-1.5em}
\paragraph{\textsc{krr} and supervised learning:}
Reasoning methods have been incorporated into supervised learning in natural language processing (\textsc{nlp})~\cite{chen2019enabling,zellers2019hellaswag} and computer vision~\cite{zellers2019recognition,aditya2019spatial,Chen_2018_CVPR} among others.  
For instance, \citeauthor{chen2019enabling} in \citeyear{chen2019enabling} used commonsense knowledge to add missing information in incomplete sentences (e.g., to expand ``pour me water'' by adding ``into a cup'' to the end); and %\citeauthor{Chen_2018_CVPR} used spatial and semantic reasoning for object recognition. 
\citeauthor{aditya2019spatial} used spatial commonsense in the Viual Question Answering (\textsc{vqa}) tasks. 
Although \textsc{lcorpp} includes components for both \textsc{krr} and supervised learning, we aim at a  \textsc{sdm} framework for robot decision-making toward achieving long-term goals, which identifies the key difference between \textsc{lcorpp} and the above-mentioned algorithms. 

\vspace{-1.5em}
\paragraph{\textsc{sdm} and supervised learning:}
Researchers have developed various algorithms for incorporating human supervision into \textsc{sdm} tasks~\cite{taylor2018improving,amiri2018multi,thomaz2006reinforcement}. Among them, \citeauthor{amiri2018multi} used probabilistic planning for \textsc{sdm} under mixed observability, where the observation model was learned from annotated datasets. 
\citeauthor{taylor2018improving} surveyed a few ways of improving robots' \textsc{sdm} capabilities with supervision (in the form of demonstration or feedback) from people. 
In comparison to the above methods, \textsc{lcorpp} is able to leverage human knowledge, frequently in declarative forms, toward more efficient and accurate \textsc{sdm}. 

% \paragraph{Learning to estimate human intention: }
% We apply our framework to the problem of identifying human intentions using their motion trajectory as the input. 
% In similar studies of this application, researchers have \citeauthor{alahi2016social} proposed the social \textsc{lstm} to consider trajectory prediction of people with interactions at large distance~\cite{alahi2016social}. 
% The developed architecture, automatically learns typical interactions that take place among trajectories which coincide in time by sharing \textsc{lstm}s of spatially proximal sequences. 
% \citeauthor{fernando2017soft+} developed a method to predict human future motion based on their trajectory and trajectory of their neighbors using \textsc{lstm} framework \cite{fernando2017soft+}. 
% None of the works involve active perception for human intention estimation. 

\vspace{.5em}
To the best of our knowledge, this is the first work on robot \textsc{sdm} that simultaneously supports supervised learning for passive perception, automated reasoning with contextual knowledge, and active information gathering toward achieving long-term goals under uncertainty.

%%%%%%%%%%%% Figure link: %%%%%%%%%%%%%%%%%%%%%%%%%%%%%%%%%%%%%
% https://docs.google.com/drawings/d/1-8qtLIt8ip5aq4-h3g6w74aug6WLI5ruTPbVcGWsBuk/edit?usp=sharing

\section{Background}
\label{sec:background}

In this section, we briefly summarize the three computational paradigms used in the buildings blocks of \textsc{lcorpp}.

\vspace{-1em}
\paragraph{\textsc{lstm}:}
Recurrent neural networks (\textsc{rnn}s) \cite{hopfield1982neural} are a kind of neural networks that use their internal state (memory) to process sequences of inputs.  
Given the input sequence vector $(x_0, x_1, ... , x_n)$, at each time-step, a hidden state is produced that results in the hidden sequence of $(h_0,h_1, ... , h_n)$. 
% The activation function of the hidden state is:
%  $$h_t = f(x_t, h_{t-1})$$
%  and in the last time-step, the output would be:
%  $$y_n=g(h_n)$$ 
% \textsc{rnn}s have been used recently in sequence predication tasks such as speech recognition~\cite{mikolov2010recurrent}, and text generation \cite{sutskever2011generating}. 
\textsc{lstm}~\cite{hochreiter1997long} network, is a type of \textsc{rnn} that includes \textsc{lstm} units. 
Each memory unit in the \textsc{lstm} hidden layer has three gates for maintaining the unit state: input gate defines what information is added to the memory unit; output gate specifies what information is used as output; and forget gate defines what information can be removed. 
\textsc{lstm}s use memory cells to resolve the problem of vanishing gradients, and are widely used in problems that require the use of long-term contextual information, e.g., speech recognition~\cite{graves2013speech} and caption generation~\cite{vinyals2015show}. 
We use \textsc{lstm}-based supervised learning for passive state estimation with streaming sensor data in this work.

\vspace{-1em}
\paragraph{P-log: }
Answer Set Prolog (\textsc{asp}) is a logic programming paradigm that is strong in non-monotonic reasoning~\cite{gelfond2014knowledge,lifschitz2016answer}. 
An \textsc{asp} program includes a set of rules, each in the form of: 
$$
	l_0 \leftarrow l_1,~ \cdots,~ l_n, ~\textnormal{not}~ l_k,~ \cdots, ~\textnormal{not}~ l_{n+k}
$$
where $l$'s are literals that represent whether a statement is true or not, and symbol \textit{not} is called default negation. The right side of a rule is the \emph{body}, and the left side is the \emph{head}. 
A rule head is true if the body is true. 

% {\color{red} a, t, and y are overloaded elsewhere. Saeid will fix the issue. }
P-log extends \textsc{asp} by allowing probabilistic rules for quantitative reasoning. 
A P-log program consists of the logical and probabilistic parts. 
The logical part inherits the syntax and semantics of \textsc{asp}. 
The probabilistic part contains \textit{pr-atoms}
%  that compute a probability for each possible world. 
% It consists of \textit{random} variables in the form of $a(X)$ where $X$ and the value of $a(X)$ ranges over finite domains. A pr-atom is:
in the form of: 
$$
pr_r (G(\eta)=w | B) =v
$$
where $G(\eta )$ is a random variable, B is a set of literals and $v \in [0,1]$.
The pr-atom states that, if $B$ holds and experiment \textit{r} is fixed, the probability of $G(\eta)=w$ is $v$.

\vspace{-1em}
\paragraph{\textsc{pomdp}s: }
Markov decision processes (\textsc{mdp}s) can be used for sequential decision-making under full observability. 
Partially observable \textsc{mdp}s (\textsc{pomdp}s)~\cite{kaelbling1998planning} generalize \textsc{mdp}s by assuming partial observability of the current state. 
A \textsc{pomdp} model is represented as a tuple $(S, A, T, R, Z, O, \gamma)$ where $S$ is the state-space, $A$ is the action set, $T$ is the state-transition function, $R$ is the reward function, $O$ is the observation function, $Z$ is the observation set and $\gamma$ is discount factor that determines the planning horizon.

An agent maintains a belief state distribution $b$ with observations ($z \in Z$) using the Bayes update rule: 
$$
	b'(s') = \frac{O(s', a, z)\sum_{s \in S} T (s,a,s')b(s)}{pr(z|a,b)}  
$$
where $s$ is the state, $a$ is the action, $pr(z|a,b)$ is a normalizer, and $z$ is an observation. 
Solving a \textsc{pomdp} produces a policy that maps the current belief state distribution to an action toward maximizing long-term utilities.

% There are representations for Logical-probabilistic inference. 
\textsc{corpp}~\cite{zhang2015corpp} uses P-log~\cite{baral2009probabilistic} for knowledge representation and reasoning, and \textsc{pomdp}s~\cite{kaelbling1998planning} for probabilistic planning.
Reasoning with a P-log program produces a set of possible worlds, and a distribution over the possible worlds. 
In line with the \textsc{corpp} work, we use P-log for reasoning purposes, while the reasoning component is not restricted to any specific declarative language.

\section{Framework}
\label{sec:framework}
In this section, we first introduce a few definitions, then focus on \textsc{lcorpp}, our robot \textsc{sdm} framework, and finally present a complete instantiation of \textsc{lcorpp} in detail. 

% We provide an overview of the framework, followed by the definitions and the algorithm of \textsc{lcorpp}. Finally,  we describe the problem of human intention estimation as one application of this work. 

% \paragraph{Overview:} Figure~\ref{fig:overview} is an illustration of the three components of \textsc{lcorpp} that includes supervised learning, automated reasoning and probabilistic planning. 

\paragraph{Definitions:} Before describing the algorithm, it is necessary to define three variable sets of $\textbf{V}^{lrn}$, $\textbf{V}^{rsn}$ and $\textbf{V}^{pln}$ that are modeled in the learning, reasoning, and planning components respectively. 
For instance, $\textbf{V}^{lrn}$ includes a finite set of variables: 
$$\textbf{V}^{lrn} = \{V^{lrn}_0,V^{lrn}_1, ...\}$$
% $$V^{rsn} = \{v^{rsn}_0,v^{rsn}_1, ..., v^{rsn}_{N^{rsn}} \}$$
% $$V^{pln} = \{v^{pln}_0,v^{pln}_1, ..., v^{pln}_{N^{pln}} \}$$

We consider factored spaces, so the three variable sets can be used to specify the three state spaces respectively, i.e., $S^{lrn},S^{rsn}$ and $S^{pln}$. 
For instance, the learning component's state space, $S^{lrn}$, can be specified by $\textbf{V}^{lrn}$, and includes a finite set of states in the form of: 
$$S^{lrn} = \{s^{lrn}_0,s^{lrn}_1, ...\}$$
% $$S^{rsn} = \{s^{rsn}_0,s^{rsn}_1, ..., s^{rsn}_{N^{rsn}} \}$$
% $$S^{pln} = \{s^{pln}_0,s^{pln}_1, ..., s^{pln}_{N^{pln}} \}$$

Building on the above definitions, we next introduce the \textsc{lcorpp} algorithm followed by a complete instantiation.

\setlength\textfloatsep{1\baselineskip}
\begin{algorithm}[t]
\caption{\textsc{lcorpp}
}
\label{alg:lcorpp}
\footnotesize
\begin{algorithmic}[1]
\REQUIRE A set of instance-label pairs $\Omega$ (training dataset), Logical-probabilistic rules $\theta$, Termination probability threshold $\epsilon$, 
\textsc{pomdp} model \text{$\mathcal{M}$}, Batch size $K$
% \Ensure State estimation $s^{pln}$
\STATE Compute classifier $\rho$ using $\Omega$ 
\STATE $C \leftarrow \textsc{cross-validate}(\rho)$, where $C$ is a confusion matrix
\STATE Update rules in $\theta$ with the probabilities in $C$

\STATE Compute action policy $\pi$ with \text{$\mathcal{M}$} (using \textsc{pomdp} solvers)
\STATE Initialize $\hat{C}$ (same shape of $C$) using uniform distributions
\STATE Initialize counter: $c \leftarrow 0$

\WHILE {$\textsc{ssub}(\hat{C} -C) > \epsilon$}
    \STATE Collect new instance $I$ \label{l1:I}
        \STATE $L \leftarrow \textsc{dt-control}(\rho,I, \text{$\mathcal{M}$})$, where $L$ is the label of $I$ \label{l1:L}
        \STATE Store instance-label pair: $\omega \leftarrow (I, L)$
        \STATE $\Omega \leftarrow \Omega \cup \omega $ 
        \STATE $c \leftarrow c+1$ \label{l1:incr}
    \IF {$c == K$} \label{l1:K}
        \STATE Compute classifier $\rho$ using augmented dataset $\Omega$ 
        \STATE $\hat{C} \leftarrow \textsc{cross-validate}(\rho)$
        \STATE Update rules in $\theta$ with the probabilities in $\hat{C}$
        \STATE $ C \leftarrow \hat{C} $
        \STATE $c \leftarrow 0$ \label{l1:0}
        \ENDIF
\ENDWHILE
\end{algorithmic}
\end{algorithm}

\subsection{Algorithms} 
We present \textsc{lcorpp} in the form of Algorithms~\ref{alg:lcorpp} and \ref{alg:dt-control}, where Algorithm~\ref{alg:lcorpp} calls the other.  

The input of \textsc{lcorpp} includes the dataset $\Omega$ for sequence classification, declarative rules $\theta$, logical facts $\beta$, and a \textsc{pomdp} model \text{$\mathcal{M}$}. 
Each sample in $\Omega$ is a matrix of size $T \times N$, where $T$ is the time length and $N$ is the number of features.
Each sample is associated with a label, where each label corresponds to state $s^{lrn} \in S^{lrn}$. 
Logical-probabilistic rules, $\theta$, are used to represent contextual knowledge from human experts. 
Facts, $\beta$, are collected at runtime, e.g., current time and location, and are used together with the rules for reasoning. 
Finally, \textsc{pomdp} model \text{$\mathcal{M}$} includes world dynamics and is used for planning under uncertainty toward active information gathering and goal accomplishments. 

Algorithm~\ref{alg:lcorpp} starts with training classifier $\rho$ using dataset $\Omega$, and confusion matrix $C$ that is generated by cross-validation.  
The probabilities in $C$ are passed to the reasoner to update probabilistic rules $\theta$. 
Action policy $\pi$ is then generated using the \textsc{pomdp} model $\mathcal{M}$ and off-the-shelf solvers from the literature.
Matrix $\hat{C}$ (of shape $C$) and counter $c$ are initialized with uniform distribution and $0$ respectively.
$\textsc{ssub}(\hat{C} -C)$ is a function that sums up the absolute values of the element-wise subtraction of matrices $\hat{C}$ and $C$.
%$$ \textsc{ssub} =  \sum_{i,j}^{|S^{lrn}|} ABS(\hat{C} -C)_{ij} $$ 
As long as the output of \textsc{ssub} is greater than the termination threshold $\epsilon$, the robot collects a new instance $I$, passes the instance along with the classifier and the model to Algorithm~\ref{alg:dt-control} to get label $L$ (Lines~\ref{l1:I}-\ref{l1:L}). 
The pair of instance and label, $\omega$, is added to the dataset $\Omega$ and the counter (c) is incremented.
If c reaches the batch size $K$, new classifier $\rho$ is trained using the augmented dataset. Then, it is cross-validated to generate $\hat{C}$, and the rules $\theta$ are updated. Also, c is set to 0 for collection of a new batch of data (Lines~\ref{l1:K}-\ref{l1:0}). 

Algorithm~\ref{alg:dt-control} requires classifier $\rho$, data instance $I$, and \textsc{pomdp} model $\mathcal{M}$  
. The classifier ($\rho$) outputs the learner's current state, $s^{lrn} \in S^{lrn}$. 
This state is then merged into $\beta$ in Line~\ref{l2:beta}, which is later used for reasoning purposes. 
A set of variables, $\hat{\textbf{V}}^{rsn}$, is constructed in Line~\ref{l2:v} to form state space $\hat{S}^{rsn}$, which is a partial state space of both $S^{rsn}$ and $S^{pln}$. 
$b^{rsn}$ is the reasoner's posterior belief. 
Belief distribution $\hat{b}^{rsn}$ over $\hat{S}^{rsn}$ bridges the gap between \textsc{lcorpp}'s reasoning and planning components: 
$\hat{b}^{rsn}$ is computed as a marginal distribution of the reasoner's output in Line~\ref{l2:mar}; and used for generating the prior distribution of $b^{pln}$ for active interactions. 
\textsc{lcorpp} initializes \textsc{pomdp} prior belief $b^{pln}$ over the state set $S^{pln}$ with $\hat{b}^{rsn}$ in Line~\ref{l2:plan}, and uses policy $\pi$ to map $b^{pln}$ to actions. 
This sense-plan-act loop continues until reaching the terminal state and the estimation label would be extracted from the reporting action (described in detail in subsection~\ref{sec:pomdp}).

\begin{algorithm}[t]\footnotesize
\caption{\textsc{dt-control}: Decision Theoretic Control 
}
\label{alg:dt-control}

\begin{algorithmic}[1]
\REQUIRE  Classifier $\rho$, Instance $I$, and \textsc{pomdp} model \text{$\mathcal{M}$}
  \STATE  Update state: $s^{lrn}\gets \mathcal{\rho}(I)$, where $s^{lrn}\in S^{lrn}$
  \STATE Collect facts $\beta$ from the world, and add $s^{lrn}$ into $\beta$ \label{l2:beta}
  \STATE $\hat{\textbf{V}}^{rsn} \leftarrow \{\hat{V}|\hat{V} \in \textbf{V}^{rsn}, \textnormal{and}~ \hat{V} \in \textbf{V}^{pln}\}$ \label{l2:v}
  \STATE Use $\hat{\textbf{V}}^{rsn}$ to form the state space of $\hat{S}^{rsn}$, where $\hat{S}^{rsn} \subset S^{rsn}$ and $\hat{S}^{rsn} \subset S^{pln}$
  \STATE Reason with $\theta$ and $\beta$ to compute belief $b^{rsn}$  over $S^{rsn}$
  \STATE Compute $\hat{b}^{rsn}$ (over $\hat{S}^{rsn}$), i.e., a marginal of $b^{rsn}$ \label{l2:mar}
  \STATE Initialize belief $b^{pln}$ (over $S^{pln}$) using $\hat{b}^{rsn}$ \label{l2:plan}
  \REPEAT
    \STATE Select action $a \gets \pi(b^{pln})$, and execute $a$
    \STATE Make observation $o$
    \STATE Update $b^{pln}$ based on $a$ and $o$ using Bayes update rule
  \UNTIL{ reaching terminal state $term\in S^{pln}$} \label{l2:end}
  \STATE $L\leftarrow \textsc{ExtractLabel}(a)$

\end{algorithmic}
\end{algorithm}

\subsection{Instantiation}

\begin{figure}[t]
  \begin{center}
    % \vspace{.5em}
    \includegraphics[width=\columnwidth]{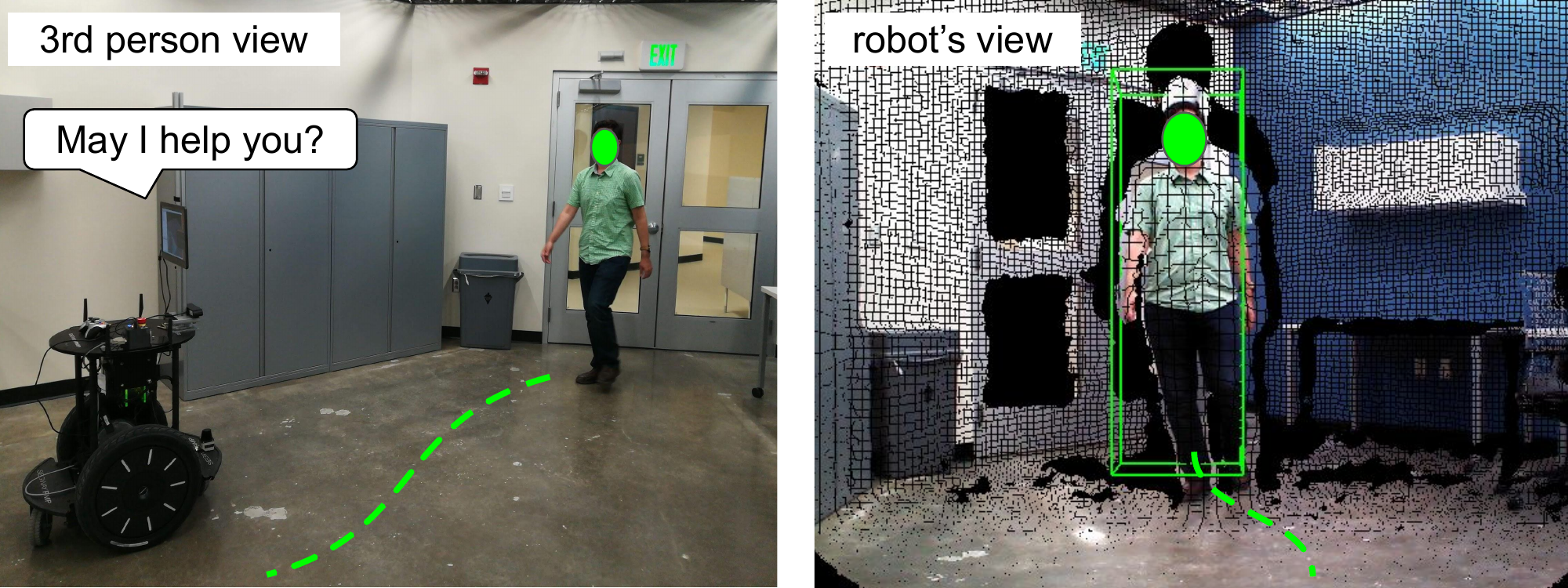}
    \vspace{-1.5em}
    \caption{Robot estimating human intention, e.g., human intending to interact or not, by analyzing human trajectories, reasoning with contextual knowledge (such as location and time), and taking human-robot interaction actions. }
    \label{fig:robot}
  \end{center}
  \vspace{-.5em}
\end{figure}

We apply our general-purpose framework to the problem of human
intention estimation using a mobile robot, as shown in Figure~\ref{fig:robot}. The robot can observe human motion trajectories using
streaming sensor data (\textsc{rgb-d} images), has contextual knowledge (e.g., visitors tend to need guidance help), and is equipped with dialog-based and motion-based interaction capabilities. The objective is to determine human intention (in our case, whether a human is interested in interaction or not).
In the following subsections, we provide technical details of a complete \textsc{lcorpp} instantiation in this domain. 
 
\subsubsection{Learning for Perception with Streaming Data}

%Robots' exteroceptive sensors are capable of perceiving various features of data over their operating period. 
In order to make correct state estimation based on the streaming sensor data while considering the dependencies at various time steps, we first train and evaluate the classifier $\rho$ using dataset $\Omega$. 
We split the dataset into training and test sets, and produce the confusion matrix $C$, which is later needed by the reasoner. 
Human intention estimation is modeled as a classification problem for the \textsc{lstm}-based learner:
$$s^{lrn} = \rho (I)$$
where robot is aiming at estimating $s^{lrn} \in S^{lrn}$ using streaming sensor data $I$. 
In our case, streaming data is in the form of people motion trajectories; and there exists only one binary variable, \textit{intention} $\in \textbf{V}^{lrn}$. 
As a result, state set $S^{lrn}$ includes only two states: 
$$
    S^{lrn}=\{s^{lrn}_0 , s^{lrn}_1 \}
$$ 
where $s^{lrn}_0$ and $s^{lrn}_1$ correspond to the person having intention of interacting with the robot or not. 
Since the human trajectories are in the form of sequence data, we use \textsc{lstm} to train a classifier for estimating human intentions with motion trajectories. 
%Accordingly, $C$ is of size $2\times 2$. 
Details of the classifier training are presented in Section~\ref{sec:exp}. Next, we explain how the classifier output is used for reasoning.

\subsubsection{Reasoning with Contextual Knowledge}
Contextual knowledge, provided by domain experts, can help the robot make better estimations. 
For instance, in the early mornings of work days, people are less likely to be interested in interacting with the robot, in comparison to the university open-house days. 
The main purpose of the reasoning component is to incorporate such contextual knowledge to help the passive state estimation. 

The knowledge base consists of logical-probabilistic rules $\theta$ in P-log~\cite{baral2009probabilistic}, a declarative language that supports the representation of (and reasoning with) both logical and probabilistic knowledge. 
The reasoning program consists of random variables set $\textbf{V}^{rsn}$. It starts collecting facts and generating $\beta$. The confusion matrix generated by the classifier's cross-validation is used to update $\theta$.
%$$\hat{V}^{lrn} = \{v | v \in V^{lrn}, v \in V^{rsn}\}$$
%$$pr(|v^{lrn} = v^{lrn})$$
The variables that are shared between the reasoning and planning components are in the set $\hat{\textbf{V}}^{rsn}$. 
The reasoner produces a belief $b^{rsn}$ over $S^{rsn}$ via reasoning with $\theta$ and $\beta$. 

In the problem of human intention estimation,  the reasoner contains random variables: 
$$ 
\textbf{V}^{rsn} = \{\texttt{\small location},~\texttt{\small time},~\texttt{\small identity},~\texttt{\small intention},\cdots\},
$$
where the range of each variable is defined as below:
\begin{align*}
&\texttt{\small location: \{classroom, library\}} \\ 
&\texttt{\small time: \{morning, afternoon, evening\}} \\
&\texttt{\small identity: \{student, professor, visitor\}} \\
&\texttt{\small intention: \{interested, not interested\}}
\end{align*}

We further include probabilistic rules into the reasoning component. 
For instance, the following two rules state that the probability of a visitor showing up in the afternoon is $0.7$, and the probability of a professor showing up in the library (instead of other places) is $0.1$, respectively. 
\begin{align*}
&\texttt{\small pr(time=afternoon|identity=visitor)=0.7.} \\
&\texttt{\small pr(location=library|identity=professor)=0.1.}
\end{align*}
% \commentsa{not sure how to make the line above smaller}

It should be noted that \emph{time} and \emph{location} are facts that are fully observable to the robot, whereas human \emph{identity} is a latent variable that must be inferred. 
Time, location, and intention are probabilistically determined by human identity. We use time and location to infer human identity, and then estimate human intention.

The  binary distribution over human intention, $\hat{b}^{rsn}$, a marginal distribution of $b^{rsn}$ over $\hat{S}^{rsn}$, is provided to the \textsc{pomdp}-based planner as informative priors.\footnote{The reasoning component can be constructed using other logical-probabilistic paradigms that build on first-order logic, such as Probabilistic Soft Logic (\textsc{psl})~\cite{bach2017hinge} and Markov Logic Network (\textsc{mln})~\cite{richardson2006markov}. 
In comparison, P-log directly takes probabilistic, declarative knowledge as the input, instead of learning weights with data, and meets our need of utilizing human knowledge in declarative forms.}

\subsubsection{Active Perception via \textsc{pomdp}-based \textsc{hri}}

\label{sec:pomdp}
% Human-Robot interaction requires sequential decision making while considering uncertainty and actions failure. 
Robots can \emph{actively} take actions to reach out to people and gather information. 
We use \textsc{pomdp}s to build probabilistic controllers. A \textsc{pomdp} model can be represented as a tuple $(S^{pln}, A, T, R, Z, O, \gamma)$. We briefly discuss how each component is used in our models: 

\begin{itemize}
\item $S^{pln}: \hat{S}^{rsn}\times S^{pln}_l \cup \{term \}$ is the state set. 
$\hat{S}^{rsn}$ includes two states representing human being interested to interact or not. 
$S^{pln}_l$ includes two states representing whether the robot has turned towards the human or not and
$term$ is the terminal state. 

\item$A: A_a \cup A_r$ is the set of actions. 
$A_a$ includes both motion-based and language-based interaction actions, including \emph{turning} (towards the human), \emph{greeting}, and \emph{moving forward} slightly. 
$A_r$ includes two actions for reporting the human being interested in interaction or not.

\item $T(s,a,s')=P(s'|s,a)$ is the transition function that accounts for uncertain or non-deterministic action outcomes where $a \in A$ and $s \in S$. 
% For instance, robot's navigation and the sensor reading can be inaccurate. 
Reporting actions deterministically lead to the \textit{term} state. 

\item $Z=\{pos, neg, none \}$ is the observation set modeling human feedback in human-robot interaction. 

\item $O(s',a,z)=P(z|a,s')$, where $z \in Z$, is the observation function, which is used for modeling people's noisy feedback to the robot's interaction actions. 

\item $R (s,a)$ is the reward function, where costs of interaction actions, $a_a \in A_a$, correspond to the completion time. A correct (wrong) estimation yields a big bonus (penalty). 
\end{itemize}

Reporting actions deterministically lead to the $term$ state. 
We use a discount factor $\gamma = 0.99$ to give the robot a long planning horizon. 
Using an off-the-shelf solver (e.g.,~\cite{kurniawati2008sarsop}), the robot can generate a behavioral policy that maps its belief state to an action toward efficiently and accurately estimating human intentions.

To summarize, the robot's \textsc{lstm}-based classifier estimates human intention based on the human trajectories. 
%Part of the structured knowledge in the reasoning component is shown in Figure~\ref{fig:problem_overview}. 
The reasoner uses human knowledge to compute a distribution on human intention. 
The reasoner's intention estimation serves as the prior of the \textsc{pomdp}-based planner, which enables the robot to actively interact with people to figure out their intention. 
The reasoning and planning components of \textsc{corpp} are constructed using human knowledge, and do not involve learning. 
The reasoning component aims at correcting and refining the \textsc{lstm}-based classifier's output, and the planning component is for active perception.

\section{Experiments}
\label{sec:exp}

In this section, we describe the testing domain (including the dataset), experiment setup, and statistical results. 

\begin{figure}[t]
  \begin{center}
    \vspace{0em}
    \includegraphics[width=0.8\columnwidth]{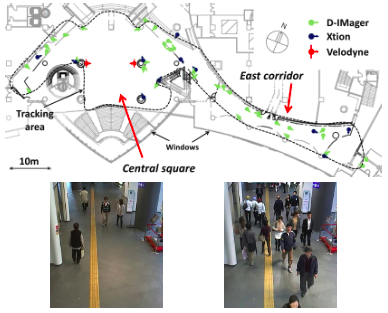}
     \vspace{-.5em}
    \caption{Tracking area and sensor setup in a shopping mall
.The dashed line shows the area covered by the sensors. The photos on right show the corridor area in the afternoon on a weekday (bottom-left) and weekend (bottom-right) [Kato \emph{et al.}, 2015]. }
	\label{fig:data}
    \end{center}
  \vspace{-1em}
\end{figure}

\subsection{Dataset and Learning Classifiers}
\label{sec:dataset}

Figure \ref{fig:data} shows the shopping center environment, where the human motion dataset~\cite{kato2015may} was collected using multiple 3D range sensors mounted overhead. 
We use the dataset to train the \textsc{lstm}-based classifier. 
Each instance in the dataset includes a human motion trajectory in 2D space, and a label of whether the human eventually interacts with the robot or not. 
There are totally 2286 instances in the dataset, where 63 are positive instances (2.7\%). 
Each trajectory includes a sequence of data fields with the sampling rate of 33 milliseconds. 
Each data field is in the form of a vector: $(x_i,~y_i,~z_i,~v_i,~\theta_{m_i},~\theta_{h_i})$. 
Index $i$ denotes the timestep. $x_i$ and $y_i$ are the coordinates in millimeter. 
$z_i$ is the human height. $v_i$ is human linear velocity in $mm/s$. $\theta_{m_i}$ is the motion angle in $radius$. $\theta_{h_i}$ is the face orientation in $radius$. 
We only use the $x$ and $y$ coordinates, because of the limitations of our robot's perception capabilities. 
%

% Features of the input vectors include the $x$ and $y$ components of human motion trajectories. 
The input vector length is 60 including 30 pairs of $x$ and $y$ values.
Our \textsc{lstm}'s hidden layer includes 50 memory units. 
In order to output binary classification results, we use a dense layer with sigmoid activation function in the output layer. 
We use Adam \cite{kingma2014adam}, a first-order gradient method, for optimization. 
The loss function was calculated using binary cross-entropy. 
For regularization, we use a dropout value of 0.2.
The memory units and the hidden states of the \textsc{lstm} are initialized to zero. 
The epoch size (number of passes over the entire training data) is 300. 
%{\color{red} Do you really mean the entire dataset? Didn't we have part of it for testing? Saeid: Changed dataset to the training data }
The batch size is 32. The data was split into sets for training (70\%) and testing (30\%).
To implement the classifier training, we used Keras \cite{chollet2015keras}, an open-source python deep-learning library.

\subsection{Illustrative Example}
\label{sec:example}

Consider an illustrative example: a \textit{visitor} to a \emph{classroom} building in the \textit{afternoon} was \emph{interested} to interact with the robot. 
The robot's goal is to identify the person's intention as efficiently and accurately as possible.

\begin{figure}[t]
  \begin{center}
    \vspace{-.5em}
    \includegraphics[width=0.98\columnwidth]{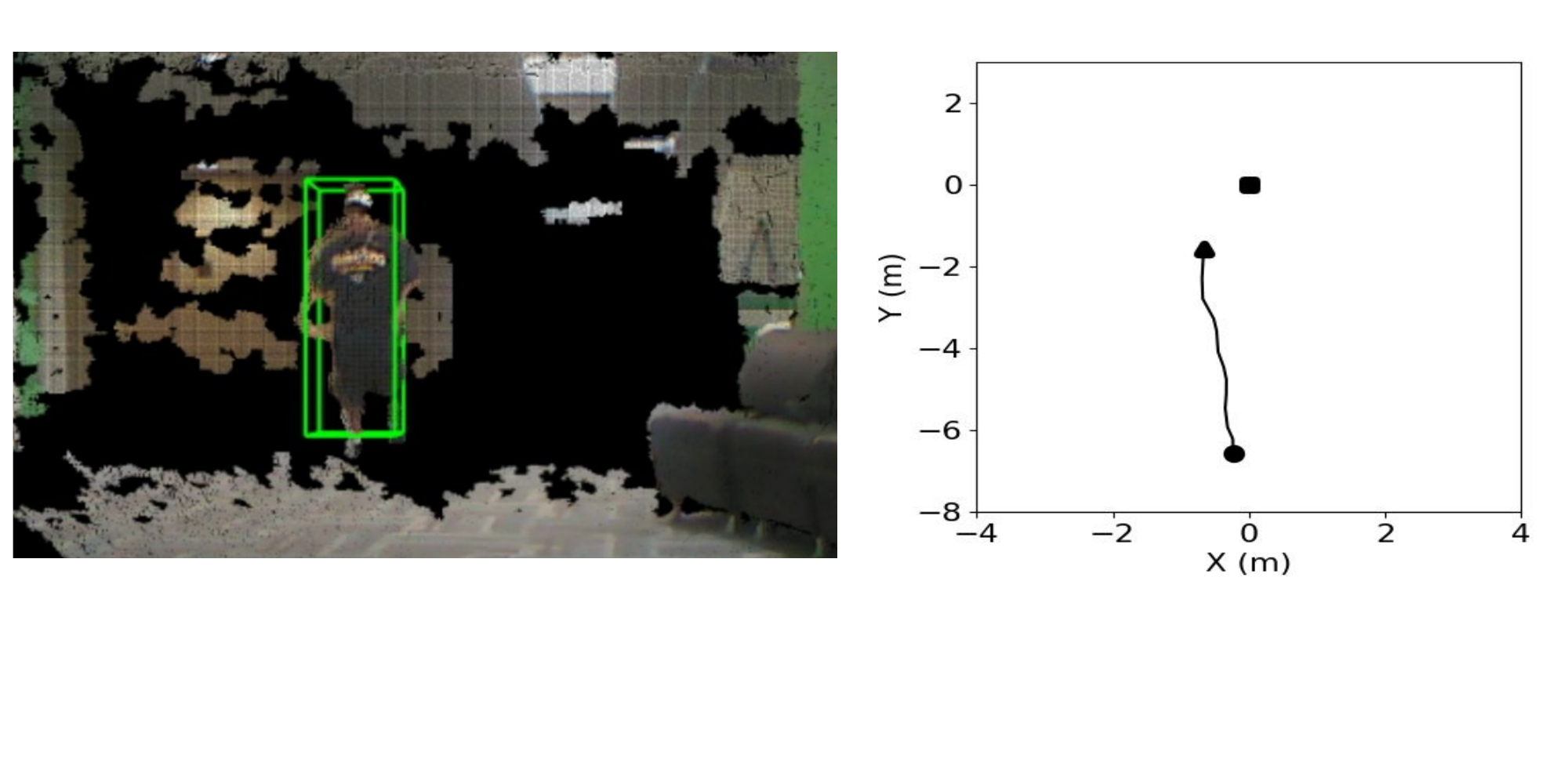}
    \vspace{-3em}
    \caption{(Left) A human detected and tracked by our Segway-based robot in the classroom building. (Right) The corresponding collected trajectory, where the robot's position is shown in square ``$\Box$'', and the human trajectory starts at dot and finishes in ``$\triangle$'' position. In this example, the human was interested in interactions.}
   \label{fig:test}
  \end{center}
  \vspace{-1em}
\end{figure}

Human motion trajectory is captured by our robot using \textsc{rgb-d} sensors. 
Figure~\ref{fig:test} presents a detected person, and the motion trajectory. 
The trajectory is passed to the \textsc{lstm}-based classifier, which outputs the person being not interested in interaction (false negative).

\begin{wrapfigure}{r}{0.38\columnwidth}
  \vspace{-2.2em}
  \begin{center}
    \includegraphics[width=0.38\columnwidth]{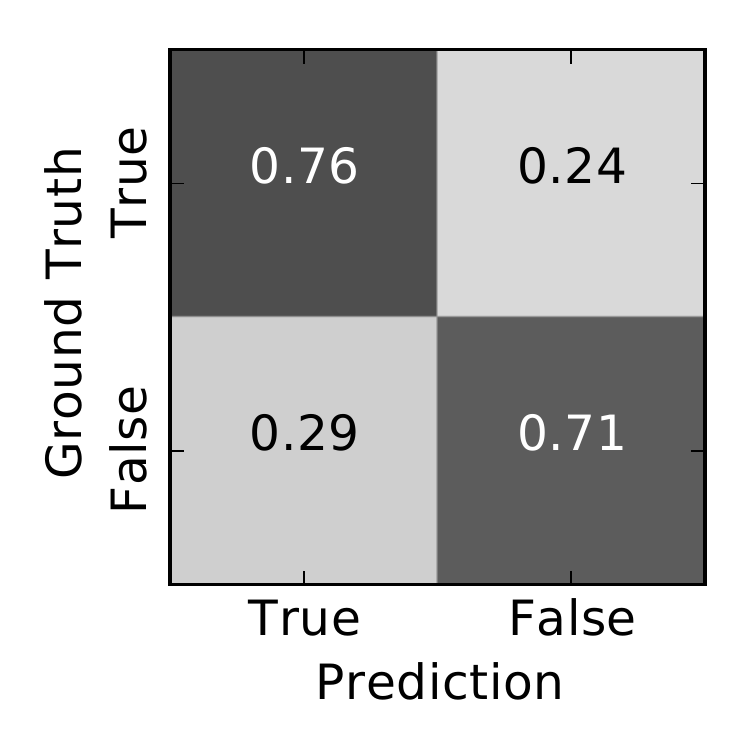}
  \end{center}
   \vspace{-2em}
  \caption{Confusion matrix of the \textsc{lstm} classifier }
\label{fig:conf_mat} 
\end{wrapfigure}

The robot then collected facts about \textit{time} and \textit{location}. 
Domain knowledge enables the robot to be aware that:  \emph{professors} and \emph{students} are more likely to show up in the classroom; and \emph{visitors} are more likely to show up in the afternoon and to interact with the robot, whereas they are less likely to be present in the classroom building.  
Also, the \textsc{lstm} classifier's confusion matrix, as shown in Figure~\ref{fig:conf_mat}, is encoded as a set of probabilistic rules in P-log, where the true-negative probability is $0.71$. 
Therefore, given all the declarative contextual knowledge, the reasoner computes the following distribution over the variable of human identity, $V^{rsn}_{id}$, in the range of [\emph{student, visitor, professor}]: 
\begin{align}
    Dist(V^{rsn}_{id}) = [0.36, 0.28, 0.36]
\end{align}
% $$pr(v^{id} =student)=0.36, pr(v^{id} =visitor)=0.28,$$
% $$pr(v^{id} =professor)=0.36$$
\setlength{\columnsep}{1.5em}%
Given the observed facts and the classifier's output, the robot queries its reasoner to estimate the distribution over possible human intentions
\begin{align}
    \hat{b}^{rsn}==[0.22, 0.78]
\end{align}
where 0.22 corresponds to the human being interested in interaction. 
$\hat{b}^{rsn}$ is the belief over state set $\hat{S}^{rsn}$ (a marginal distribution of both $S^{rsn}$ and $S^{pln}$).

The reasoner's output of $\hat{b}^{rsn}$ is used for initializing the belief distribution, $b^{pln}$, for the \textsc{pomdp}-based planner:
$$b^{pln}=[0.22,0.78,0,0,0]$$ 
where $b^{pln}$ is over the state set of $S^{pln}$ as described in Section~\ref{sec:pomdp}. 
For instance, $s_0^{pln}\in S^{pln}$ is the state where the robot has not taken the ``turn'' action, and the human is interested in interaction. 
Similarly, $s_3^{pln}\in S^{pln}$ is the state where the robot has taken the action ``turn'', and the human is not interested in interaction.

\begin{figure}[ht]
  \begin{center}
    %  \vspace{.5em}
    \includegraphics[width=0.98\columnwidth]{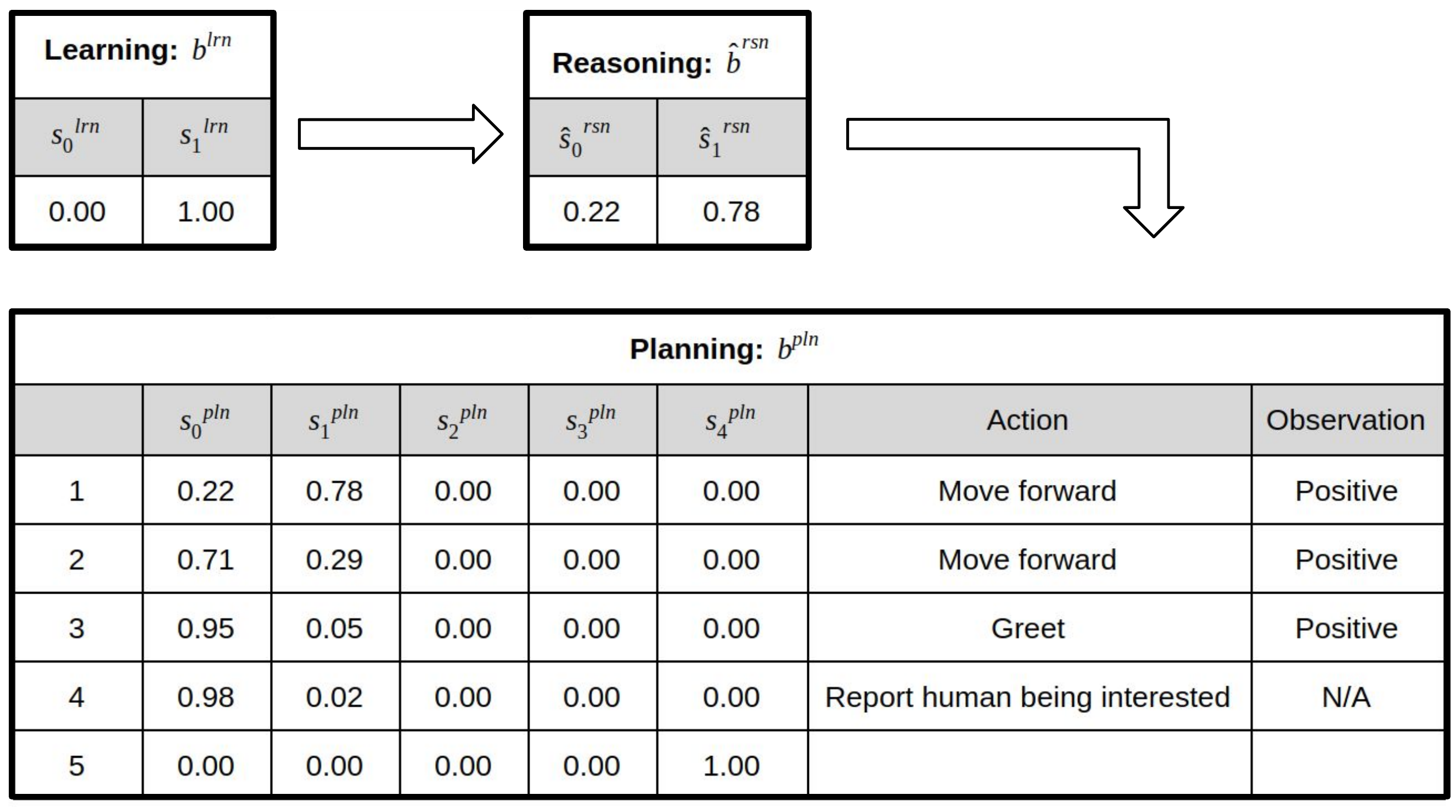}
    \vspace{-.2em}
    \caption{An illustrative example of information flow in \textsc{lcorpp}: the learning component generates ``facts'' and probabilistic rules for the reasoner, and the reasoner computes a prior distribution for the planner, which actively interacts with the environment. }
    \label{fig:illustrative}
  \end{center}
  \vspace{-1em}
\end{figure}
% https://docs.google.com/drawings/d/1hTFHNyxQdKj8_Ot1EyeILmPi-fviWCCY4DzjOAZPvjw/edit?usp=sharing

During the action-observation cycles (in simulation), policy $\pi$ maps $b_{pln}$ to ``greet'' and ``move forward'' actions, and $b^{pln}$ is updated until the robot correctly reported human intention and reached terminal state ($s_4^{pln}$). 
The actions, corresponding human feedback, and belief update are presented in Figure~\ref{fig:illustrative}. 
Although, the \textsc{lstm} classifier made a wrong estimation, the reasoner and planner helped the robot successfully recover from the wrong estimation. 

\subsection{Experimental Results}
\label{sec:results}
We did pairwise comparisons between \textsc{lcorpp} with the following methods for human intention estimation to investigate several hypotheses. Our baselines include:
\textbf{Learning (L)}: learned classifier only. 
\textbf{Reasoning (R)}: reasoning with contextual knowledge. 
\textbf{Planning (P)}: \textsc{pomdp}-based interaction with uniform priors. 
\textbf{Learning+Reasoning (L+R)}: reasoning with the classifier's output and knowledge. 
\textbf{Reasoning+Planning (R+P)}: reasoning with knowledge and planning with \textsc{pomdp}s. 
%\textbf{Ours}: our approach. 

Our experiments are designed to evaluate the following hypotheses. I) Given that \textsc{lcorpp}'s prior belief is generated using reasoner and learner, it  would outperform baselines in intention estimation in F1 score while receiving less action costs; II) In case of inaccurate knowledge, or  III) the learner not having a complete sensor input, \textsc{lcorpp}'s planner can compensate these imperfections by taking more actions to maintain higher F1 score. IV) In case of scarcity of data, robot's most recent interaction can be used to augment the dataset and improve overall performance of \textsc{lcorpp}.      

In each simulated trial, we first sample human identity randomly, and then sample time and location accordingly, using contextual knowledge, such as professors tend to showing up early. 
According to the time, location, and identity, we sample human intention. 
Finally, we sample a trajectory from the test set of the dataset, according to the previously sampled human intention. 
We added $30\%$ noise to the human reactions (robot's observation) being compliant with the ground truth but independent from robot's actions. 
\textsc{lcorpp} requires considerable computation time for training the classifier ($\sim$10 min) and generating the
\textsc{pomdp}-based action policy ($\sim$1 min). 
The training and policy generation are conducted offline, so they do not affect the runtime efficiency. 
Reasoning occurs at runtime, and typically requires less than 1 millisecond.

\begin{figure}[t]
  \begin{center}
    % \vspace{.5em}
    \includegraphics[width=\columnwidth]{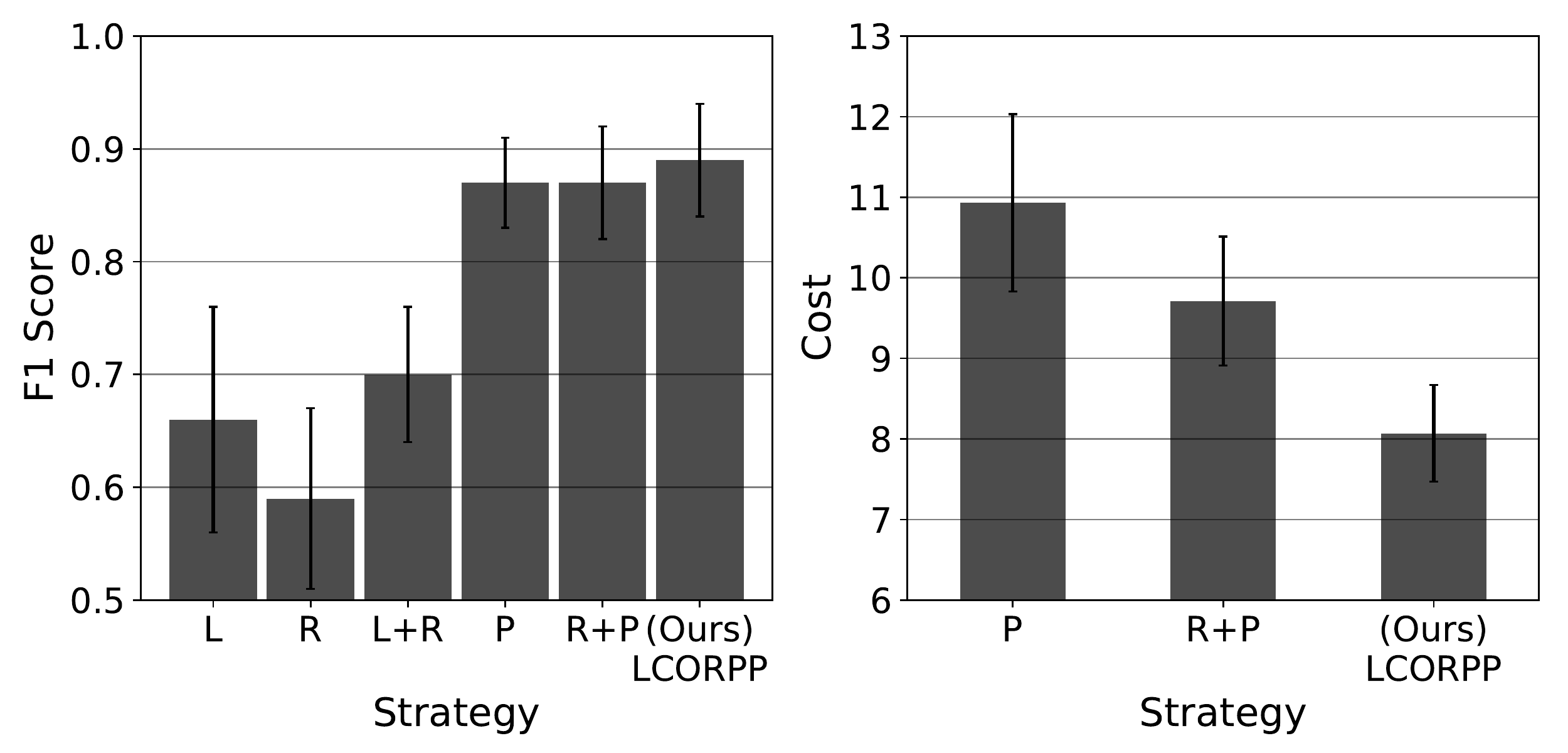}
    \vspace{-1em}
    \caption{Pairwise comparisons of \textsc{lcorpp} with five baseline sequential decision-making strategies. The right subfigure excludes the strategies that do not support active human-robot interaction and hence produce zero costs.}
    \label{fig:exp_1}
  \end{center}
  \vspace{-.5em}
\end{figure}

\vspace{-1em}
\paragraph{\textsc{lcorpp} vs. Five Baselines:}
Figure~\ref{fig:exp_1} shows the overall comparisons using the six \textsc{sdm} strategies, each number is an average of 5000 trials, where the setting is the same in all following experiments. 
The three strategies, \textbf{P}, \textbf{R+P} and \textsc{lcorpp}, include the \textsc{pomdp}-based planning component, and perform better than no-planning baselines in F1 score. 
 Among the planning strategies, ours produces the best overall performance in F1 score, while reducing the interaction costs (dialog-based and motion-based)(Hypothesis I). 

\vspace{-1em}
\paragraph{Inaccurate Knowledge:}
In this experiment, we evaluate the robustness of \textsc{lcorpp} to inaccurate knowledge (Hypothesis II). 
Our hypothesis is that, in case of contextual knowledge being inaccurate, \textsc{lcorpp} is capable of recovering via actively interacting with people. 
We used knowledge bases (\textsc{kb}s) of different accuracy levels: {\bf High}, {\bf Medium}, and {\bf Low}. 
% Informally, knowledge bases at the three levels are `useful', `useless', and `misleading' respectively .
A high-accuracy \textsc{kb} corresponds to the ground truth. 
Medium- and low-accuracy \textsc{kb}s are incomplete, and misleading respectively. 
For instance, low-accuracy knowledge suggests that \emph{professors} are more likely to interact with the robot, whereas \emph{visitors} are not, which is opposite to the ground truth. 
\begin{figure}[t]
  \begin{center}
    % \vspace{.5em}
    \includegraphics[width=\columnwidth]{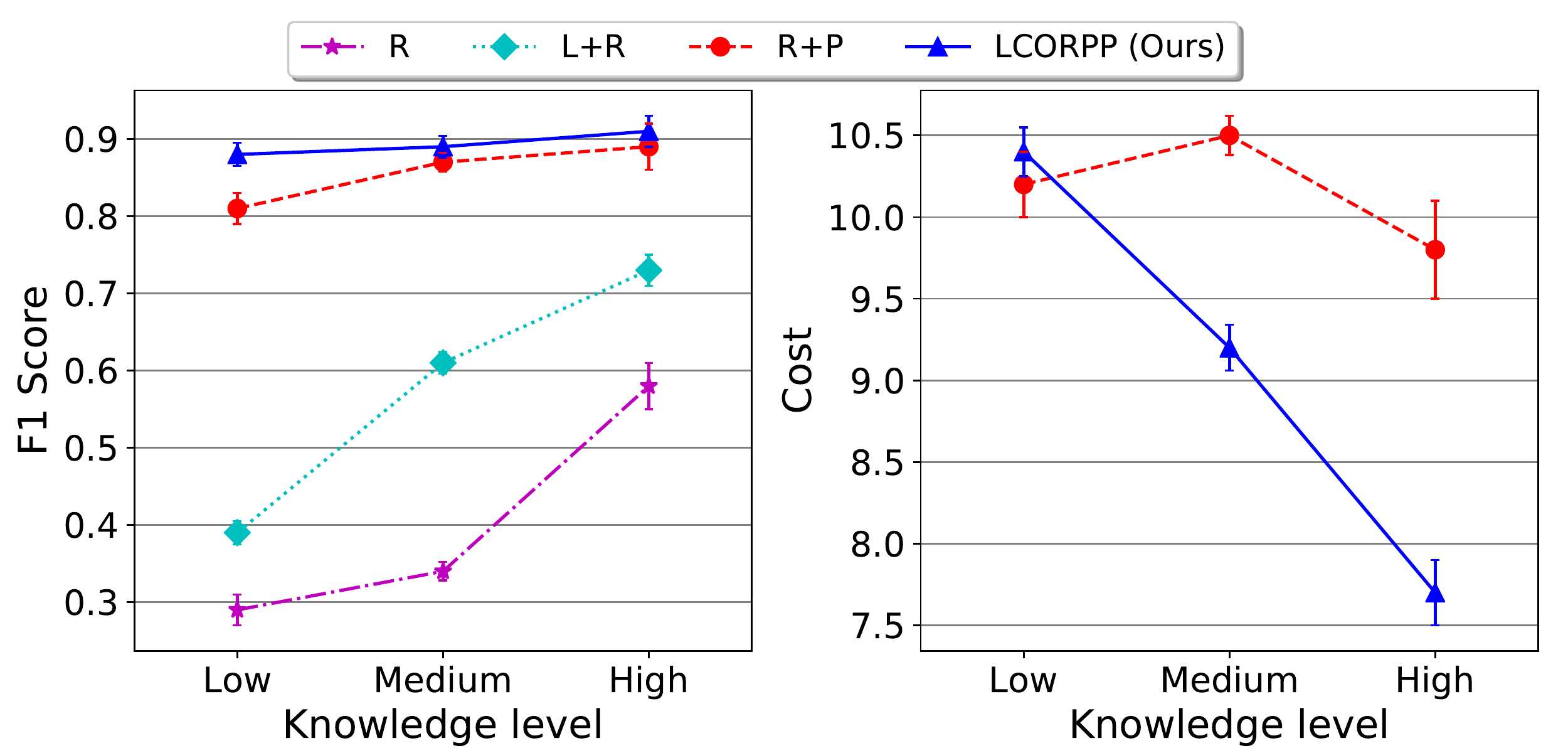}
    \vspace{-1em}
    \caption{Performances of \textsc{lcorpp} and baselines given contextual knowledge of different accuracy levels: High, Medium and Low. Baselines that do not support reasoning with human knowledge are not included in this experiment.}
    \label{fig:exp_2}
  \end{center}
  \vspace{-.5em}
\end{figure}

Figure~\ref{fig:exp_2} shows the results where 
we see the performances of \textbf{R} and \textbf{L+R} baseline strategies drop to lower than 0.4 in F1 score, when the contextual knowledge is of low accuracy. 
In F1 score, neither \textbf{R+P} nor \textsc{lcorpp} is sensitive to low-accuracy knowledge, while \textsc{lcorpp} performs consistently better than \textbf{R+P}. 
In particular, when the knowledge is of low accuracy, \textsc{lcorpp} retains the high F1 score (whereas \textbf{R+P} could not) due to its learning component.  

Additional experimental results on Hypotheses III and IV are provided in the supplementary document.

\vspace{-0.6 em}
\section{Conclusions \& Future Work} 
\label{sec:conclusion}

In this work, we develop a robot sequential decision-making framework that integrates supervised learning for passive state estimation, automated reasoning for incorporating declarative contextual knowledge, and probabilistic planning for active perception and task completions. 
The developed framework has been applied to a human intention estimation problem using a mobile robot. 
Results suggest that the integration of supervised deep learning, logical-probabilistic reasoning, and probabilistic planning enables simultaneous passive and active state estimation, producing the best performance in estimating human intentions. 

This work enables research directions for future work. 
It can be interesting to consider situations where contradictions exist among \textsc{lcorpp}'s components, e.g., the classifier's output consistently does not agree with the reasoning results from human knowledge. 
Additionally, we can leverage multi-modal perception for cases where people approach the robot outside its camera sensor range. 
When the robot has to estimate the intentions of multiple humans, more advanced human-robot interaction methods can be deployed. 

\section*{Acknowledgments}
\footnotesize
This work has taken place in the Autonomous Intelligent Robotics (AIR)
Group at SUNY Binghamton. AIR research is supported in part by grants from the National Science Foundation (IIS-1925044), Ford Motor Company, and SUNY Research Foundation.

{\footnotesize
\bibliographystyle{aaai}
\bibliography{references}
}
%\end{document}
%%%%%%%%%%%%%%%%%%%%%%%%%%%%%%%%%%%%%%%% APPENDIX %%%%%%%%%%%%%%%%%%%%%%%%%%%%%%%%%%%%%%%%%%%%%%%%%%%%%%%%%%%%%%%%%%%%%%%%

\cleardoublepage

% command to correct it. You may not alter the value below 2.5 in
% \title{Learning and Reasoning for Robot Sequential Decision Making under Uncertainty\\
\section*{Supplementary Materials}
%Your title must be in mixed case, not sentence case. 
% That means all verbs (including short verbs like be, is, using,and go), 
% nouns, adverbs, adjectives should be capitalized, including both words in hyphenated terms, while
% articles, conjunctions, and prepositions are lower case unless they
% directly follow a colon or long dash

\maketitle

In this supplementary document, we present additional experimental results, and technical details of \textsc{lcorpp}'s implementation on a mobile robot platform. 

\subsection*{Partial Motion Trajectories}
\label{sec:results}

Within the human intention estimation context, the robot's goal is to identify human intention as accurately and early as possible. 
However, there is the trade-off between efficiency and accurate. 
For instance, reaching out to people early usually means the robot only has a short piece of human motion trajectory, and frequently produces less accurate intention estimation result. 
In this experiment, we evaluate how sensitive our platform is to partial human motion trajectories (Hypothesis III). 
The experimental results can provide robot practitioners a reference on how early a modern robot can produce reasonably accurate intention estimation results.

\begin{figure}[h]
  \begin{center}
    % \vspace{.5em}
    \includegraphics[width=0.8\columnwidth]{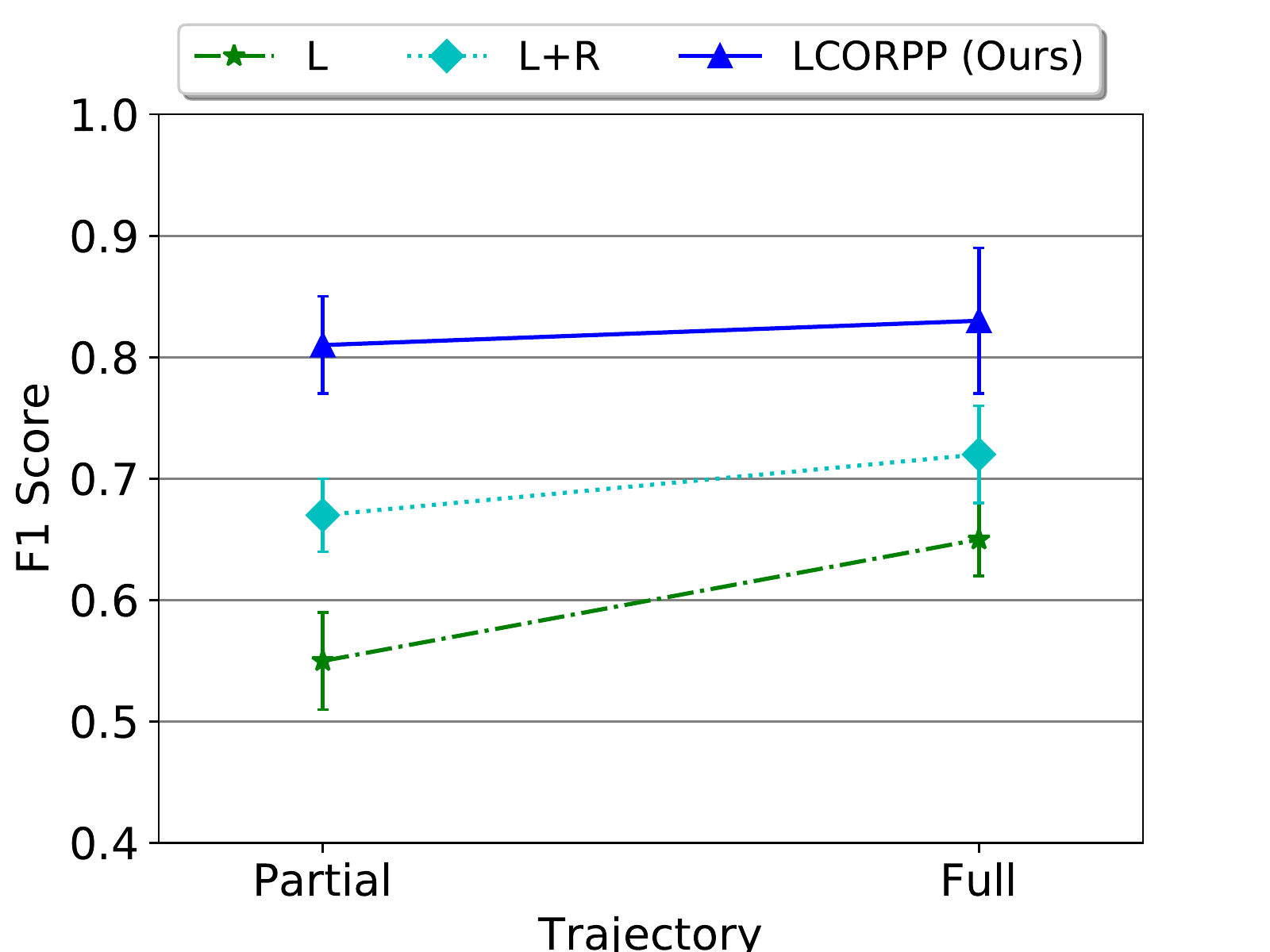}
    % \vspace{-.5em}
    \caption{Experiments with incomplete sensor input (trajectory), in comparison to the two ``learning-included'' baselines. }
    \label{fig:partialsensor}
  \end{center}
  \vspace{-1em}
\end{figure}

We only provided part of each trajectory to the classifier (the first quarter in our case). 
The goal is to evaluate, if the robot is forced to produce the estimation result using only $1/4$ piece of the trajectories, how reliable the three ``learning-included'' strategies are, where ``L'' includes only the \textsc{lstm}-based classifiers, ``L+R'' further combines the reasoner, and \textsc{lcorpp} is our approach. 

The results are presented in Figure~\ref{fig:partialsensor}. 
We can see, in case of partial trajectories, \textsc{lcorpp} is able to retain its human intention estimation accuracy in F1 score. 
We attribute this achievement to the interaction capability of \textsc{lcorpp} based on experimental results (Figure 8) shown in the main body of the paper.

\begin{figure}[h]
  \begin{center}
    % \vspace{.5em}
    \includegraphics[width=0.8\columnwidth]{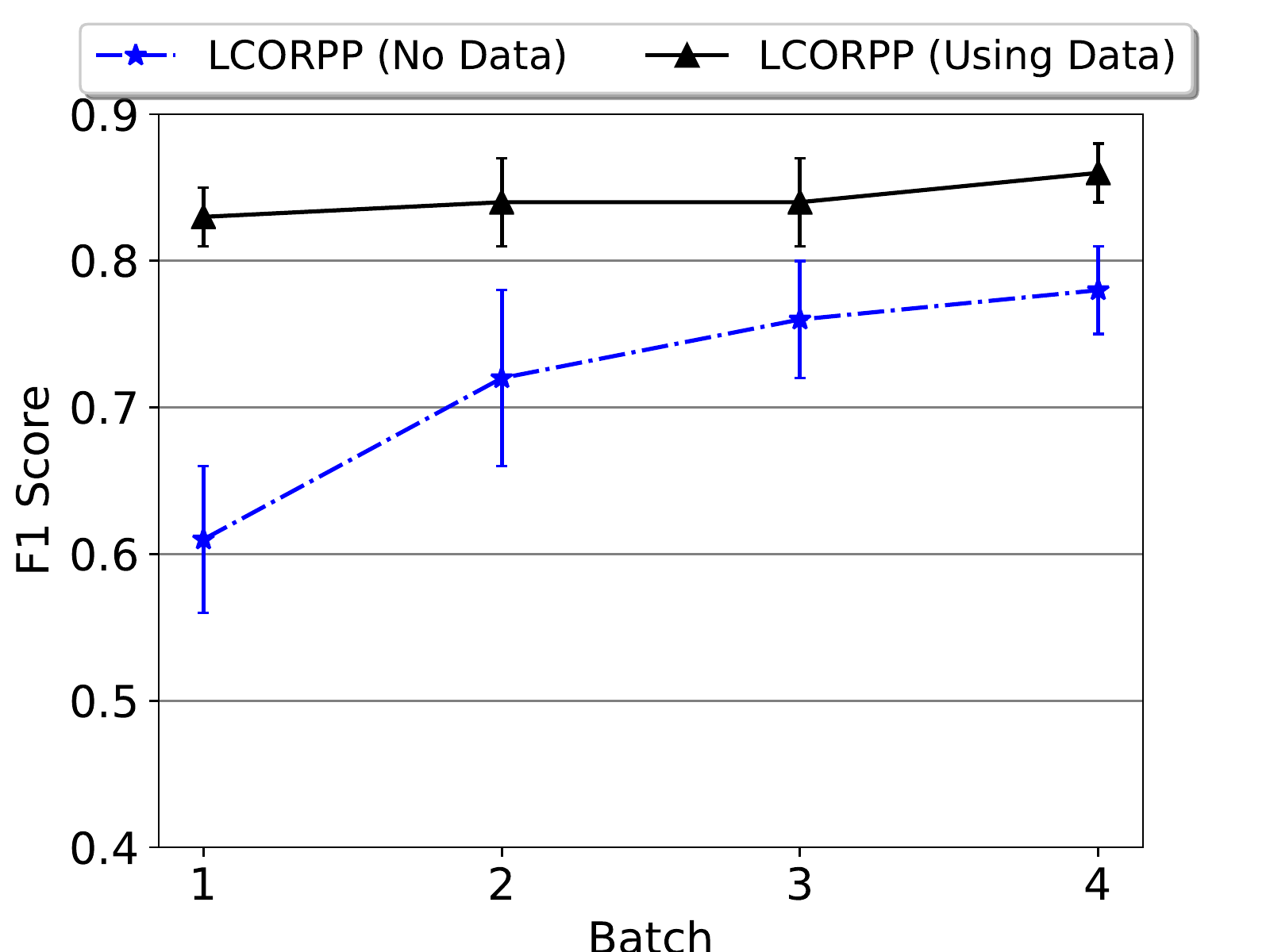}
    % \vspace{-.5em}
    \caption{Active data augmentation from the experience of human-robot interaction. }
    \label{fig:batches}
  \end{center}
  \vspace{-1em}
\end{figure}

\subsection*{Active Data Augmentation}

In this experiment, we explored the effectiveness of augmenting the dataset (for supervised learning) using the experience of human-robot interaction, including trajectories and corresponding labels generated by the task planner. 
Our hypothesis is that, as more trajectories are collected and added into the dataset, our robot is able to re-train the classifier, and produce increasingly better overall performance in F1 score (Hypothesis IV). 

Instead of training the initial classifier using all data, we divided the dataset into four batches, each containing 500 instances. 
In the first batch, we ran \textsc{lcorpp} without offering any training data. 
As a result, the ``supervised learning'' component produced a random classifier. 
In the subsequent batches, robot interacts with human using the interaction instances collected from early bathes. 

Figure~\ref{fig:batches} shows the results. 
As the robot becomes more experienced (i.e., more instances of human-robot interaction trials become available), \textsc{lcorpp} performs better in terms of F1 score.

\subsection*{Technical Details in Robot Implementation}

We use a Segway-based mobile robot platform, RMP110, for experimental trials in the real world. 
The platform is equipped with SICK laser sensor for mapping, localization and navigation. 
Astra Orbecc RGB-D camera is used for human detection and interaction. 
The robot's software runs on Robot Operating System (ROS)~\cite{quigley2009ros}. 
Our implementation is established on the Building Wide Intelligence codebase that is available to the public on Github~\cite{khandelwal2017bwibots}. 

For active information collection, the robot has to take actions while receiving (local and unreliable) observations. 
We manually provide the robot its initial position for localization. 
After that, to \texttt{move forward}, the robot navigates to a goal position by a fixed distance toward to the human. 
To \texttt{greet}, the robot uses the ``sound play'' package in ROS to convert pre-defined text for greeting into audio. 
To observe human's vocal feedback, we use the \texttt{pocketsphinx} package~\cite{huggins2006pocketsphinx} for speech recognition. 
A set of pre-defined positive utterances indicate human's eagerness to interact.

\begin{figure}[t]
  \begin{center}
    % \vspace{.5em}
    \includegraphics[width=0.8\columnwidth,height=0.9\columnwidth]{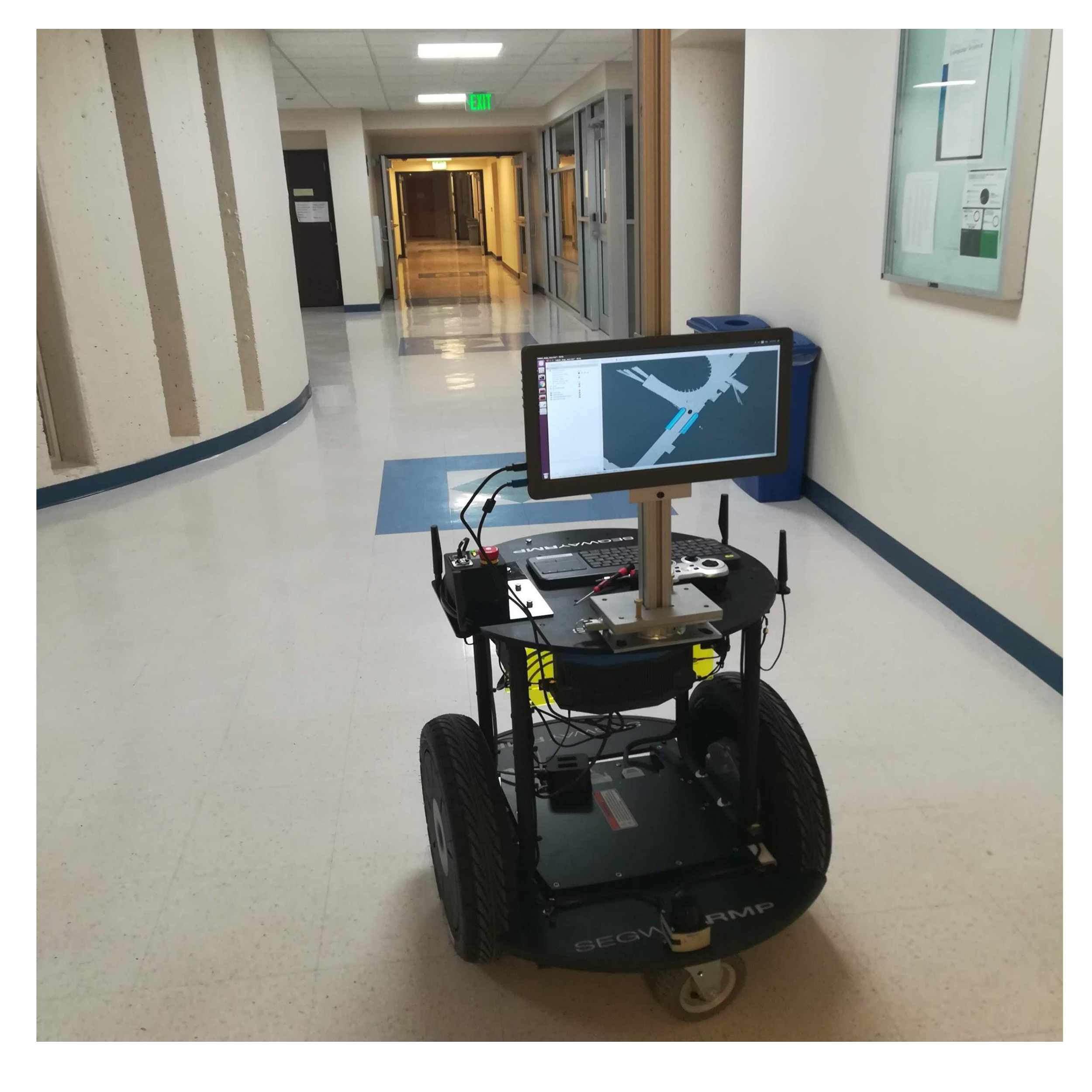}
    \vspace{-.5em}
    \caption{RMP110, the Segway-based mobile robot platform used for experimental trials in the real world. }
    \label{fig:segbot}
  \end{center}
  \vspace{-.5em}
\end{figure}

\end{document}